\documentclass[a4paper,tikz]{article}

\usepackage{INTERSPEECH2019}
\usepackage{hyperref}
\usepackage{multirow}
\usepackage[table,xcdraw]{xcolor}
\usepackage{verbatim}
\usepackage{balance}

\title{Modeling user context for valence prediction from narratives }
\name{Aniruddha Tammewar$^1$, Alessandra Cervone$^1$, Eva-Maria Messner$^2$, Giuseppe Riccardi$^1$}
\address{
  $^1$Signals and Interactive Systems Lab, University of Trento\\
  $^2$Clinical Psychology and Psychotherapy, University of Ulm}
\email{\{aniruddha.tammewar, alessandra.cervone, giuseppe.riccardi\}@unitn.it, eva-maria.messner@uni-ulm.de}

\begin{document}

\maketitle
\begin{abstract}
Automated prediction of valence, one key feature of a person's emotional state, from individuals' personal narratives may provide crucial information for mental healthcare (e.g. early diagnosis of mental diseases, supervision of disease course, etc.). 
In the Interspeech 2018 ComParE Self-Assessed Affect challenge, the task of valence prediction was framed as a three-class classification problem using 8 seconds fragments from individuals' narratives. 
As such, the task did not allow for exploring contextual information of the narratives. 
In this work, we investigate the intrinsic information from multiple narratives recounted by the same individual in order to predict their current state-of-mind. 
Furthermore, with generalizability in mind, we decided to focus our experiments exclusively on textual information as the public availability of audio narratives is limited compared to text.
Our hypothesis is that context modeling might provide insights about emotion triggering concepts (e.g. events, people, places) mentioned in the narratives that are linked to an individual's state of mind.
We explore multiple machine learning techniques to model narratives. 
We find that the models are able to capture inter-individual differences, leading to more accurate predictions of an individual's emotional state, as compared to single narratives.
\end{abstract}

\section{Introduction}

The recollection and novel interpretation of personal narratives is a key feature of psychotherapeutic approaches \cite{vromans2011narrative}. The use of narratives in psychotherapy is rooted in the association between mood and recollection of episodic memories \cite{sumner2010overgeneral}. Earlier work showed an interrelation between personal storytelling and self-reported affect (mood) as well as mental health and word use in personal narratives \cite{rathner2018state, rathner2018did}. 

This work investigates the possibility of automatically predicting individuals' self-reported affect using the context of multiple narratives recounted by the same subject. As such, our approach could be the first step towards automatized personal narrative analysis to assess individuals' affective state. Software for automatized narrative analysis could prove useful in several applications, including detection of mood or mental disease, 
distribution of tailored internet-and mobile-based interventions and evaluation of therapy outcome \cite{rathner2018state}.

Previous research on affect relied on self-report resulting in susceptibility to subjectivity and socially desirable answer behavior \cite{steenkamp2010socially}. Given informed consent of individuals, the automatized analysis of written personal narratives could be used on a broad scale in the future. This could especially benefit longitudinal data assessments because individuals tend to be less compliant with study protocols over extended time periods \cite{donkin2012motivators}. Currently, a vast amount of text data is easily accessible online, resulting in the potential to minimize the amount of active user input required to monitor affective states \cite{markowetz2014psycho}. Furthermore, automatized text analysis could shed light on concepts or entities that are connected to mood and therefore give individuals, insight into their personal triggers for positive or negative emotions. 

Given the major impact on the mental and physical health of individual's affective state-of-mind \cite{houben2015relation}, state-of-mind prediction (via \textit{valence} scores \cite{russell2003core, russell2012sinuosity}) was the focus of the Self-Assessed Affect Subchallenge, part of the Interspeech 2018 Computational Paralinguistics Challenge (ComParE) \cite{schuller2018interspeech}. This challenge utilized data from the Ulm State-of-Mind corpus (USoMS) \cite{rathner2018state}, a dataset of spoken personal narratives recollected by individuals (4 narratives per individual). 
Instead of the full narratives, attendees were provided with 8 seconds fragments of the current narrative (the one uttered just before the valence score to be predicted). 
Thus, participants \cite{montacie2018vocalic,syed2018computational,gorrostieta2018attention} did not have access to the full narratives recounted by individuals. 

Our work is based on the hypothesis that the context provided by the full history of both current and previous narratives recounted by the same individual might be useful for the task of valence prediction. In particular, we argue that modeling multiple narratives by the same subject could not only be helpful for predicting state-of-mind, but might also provide interesting insights about emotion-triggering concepts (e.g. events, people) and other features beyond direct manifestation through lexicalization ( e.g. sad, happy , etc.) which might be associated to self-reported affect for individuals. 
Moreover, we investigate the possibility of utilizing solely the textual information in our experiments, in order to verify the applicability of the approach for cases where the acoustics might not be available. In order to test our hypothesis, we train different machine learning (ML) models with and without the previous context.

The structure of the paper is the following: first, we provide details about the data used (Section \ref{sec:corpus}), then in Section \ref{sec:methodology} we describe our methodology describing the ML models used and preprocessing of the data. Afterwards, we report the results of our experiments which show the importance of modeling context across different ML models. In Section \ref{sect:disc} we provide a qualitative analysis of the models and find that they seem to capture relevant aspects of individuals state of mind. In Section \ref{sect:conc} we draw the conclusions of our work.

\section{The Ulm State of Mind dataset}
\label{sec:corpus}
\begin{figure}[]
  \includegraphics[width=8.0cm]{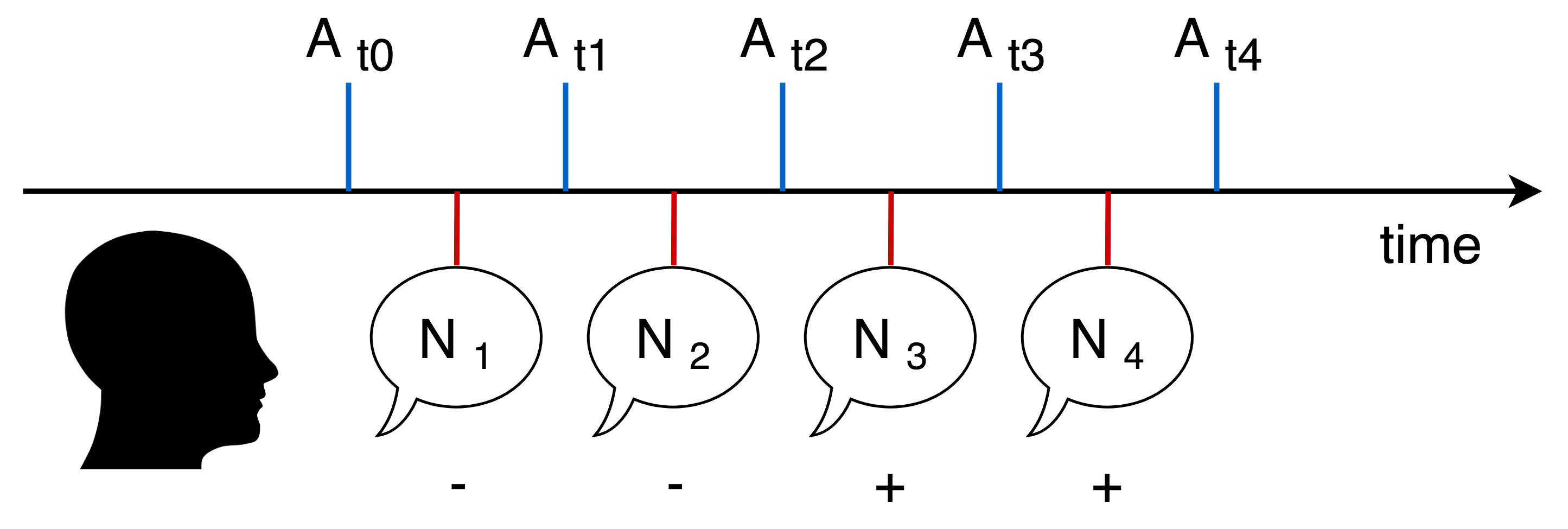}
  \caption{Data collection process in the Ulm State of Mind corpus: participants were asked to self-report their affect ($A_{t0}$), then recount a negative narrative ($N_{1}$,$-$), report their affect ($A_{t1}$), recount another negative narrative ($N_{2}$,$-$), report their affect ($A_{t2}$), recount a positive narrative ($N_{3}$,$+$), report their affect ($A_{t3}$), recount a positive narrative  ($N_{4}$,$+$) and report their affect one final time ($A_{t4}$).}
  \label{fig:corpus}
  \vspace{-0.3cm}
\end{figure}
The Ulm State of Mind corpus (USoMS) \cite{schuller2018interspeech} is a dataset of personal narratives with self-reported affect information.
A part of USoMS was used in the INTERSPEECH 2018 Computational Paralinguistics Challenge \cite{schuller2018interspeech}. This subset of the original dataset was called Self-Assessed Affect Sub-Challenge and consists of 100 speakers (85 f, 15 m, age 18-36 years, mean 22.3 years, std. dev. 3.6 years). Individuals reported two negative and two positive personal narratives, each for 5 minutes, and assessed their affect before and after each narrative on a 10 point Likert scale (see Figure \ref{fig:corpus}). Affect was collected using the affect grid \cite{russell2003core} on the two independent domains arousal (spanning from sleepy to excited) and valence (spanning from negative to positive). The resulting files were transcribed manually. For this paper, the self-reported valence scores were grouped into Low (0-4) Medium (5-7) and High (8-10) as in last year´s sub-challenge.


\section{Methodology}
\label{sec:methodology}
Contextual information such as previous narratives of the same subject, its valence state before uttering the narrative and other user features may contain information crucial for identifying the user's current valence state. We try to incorporate such information in our experiments through feature engineering, various machine learning models and DNN architectures\footnote{We plan to make our code available at \url{https://tinyurl.com/som-context}}.

\subsection{Features}
\label{sect:fatures}
In the experiments we use different combinations utilizing one or more of the following features:\\
\textbf{Sentiment polarity score (pol):} provided by the `Sentiment Analyzers' module from textblob-de \cite{loria2014textblob} for a narrative. \\
\textbf{Word embeddings (word embs):} GloVe \cite{pennington2014glove} word embeddings (dimension:300) for German, pretrained on WikiPedia.\\
\textbf{Tf-idf:} tf-idf \cite{sparck1972statistical} feature vector for a narrative. The vectorizer is trained on training data using scikit--learn \cite{scikit-learn}. We varied ngram range and found that combination of unigrams and bigrams gives best results. In all the experiments presented, the tf-idf vectorizer uses the same ngram range.\\
\textbf{Previous valence class (prev\_val):} for a given narrative, it is the valence state of the user before starting the narrative. Compared to previous features, which are automatically extracted from text, in our corpus this is a gold feature since we have the true labels of the previously self-reported affect, rather than the predicted ones. 
This condition, however, may not hold in real-world scenarios where users might not be asked to report their affect after each narrative.
\subsection{Models}
\label{sect:models}
We explore both neural and classical ML algorithms to model contextual information for valence prediction.
\subsubsection{Linear SVM}
\label{sect:svm}
We experimented with various classic Machine Learning algorithms including XGBoost, Support Vector Machine (SVM) with different kernels, and found that Linear SVM performs well for our problem. We experiment with creating document vectors with two methodologies,
 \textit{`word embeddings'} and \textit{`tf-idf'}. While in the first method we take the average of word embeddings of all the words present in the narrative, to generate tf-idf vectors we train a vectorizer on narratives from the training data. In the vectorizer, we use l2 normalization, remove stopwords and use ngram range of (1,2).
 
For example, in the case of valence prediction of a narrative in isolation, we use the feature vector produced for that narrative as an input for the classifier. In order to add other features (see section \ref{sect:fatures}), we simply concatenate document (narrative in our case) vectors with other narrative level features such as polarity score, previous valence score, etc. Furthermore, to integrate context in this model, we create narrative level features for other context narratives in a similar fashion and concatenate all the vectors, to form the final input vector for SVM. We discuss more about the specific experiments in Section \ref{sect:results}
 
 \subsubsection{DNN architectures}
 \label{sect:dnn}

\begin{figure}[]
\begin{center}
    \includegraphics[width=5.7cm]{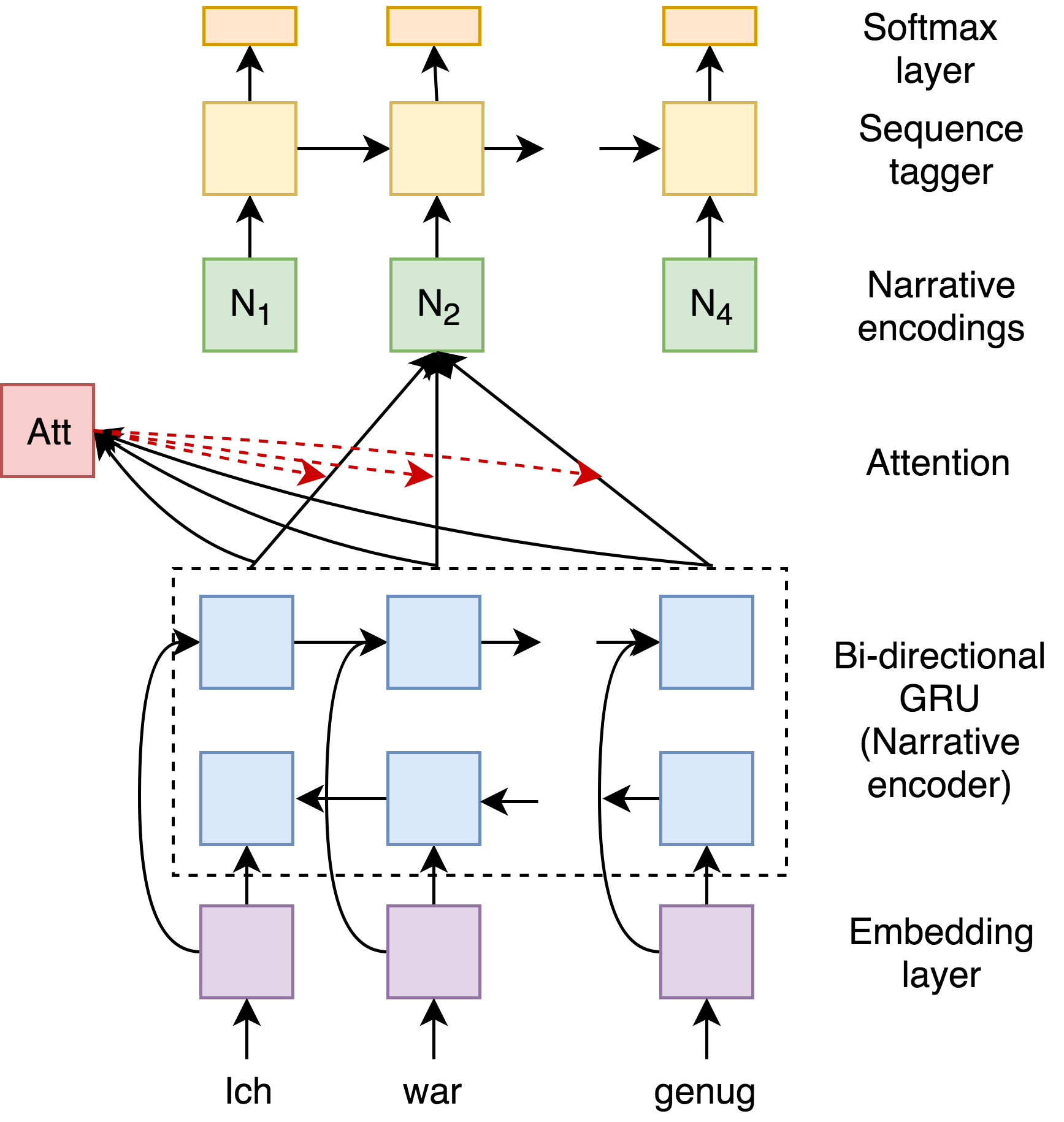}
  \caption{DNN based sequence tagging architecture. The lower part uses bidirectional RNN with attention to generate narrative encodings. These encodings are fed as inputs to the unidirectional RNN on top to predict valence class for each narrative.}
  \label{fig:architecture}
  \vspace{-0.7cm}
  \end{center}
\end{figure}

 Similar to our approach using SVM we first experiment with a narrative in isolation, to set up a baseline. For this task, we use Bidirectional recurrent neural networks \cite{schuster1997bidirectional} (with Gated Recurrant Unit \cite{cho2014learning} cells) architecture with attention \cite{bahdanau2014neural} as a multi-class classifier to predict the valence class. To integrate context, we use two different architectures with a slight difference, based on the amount of context to be encoded.\\
\textbf{Sequence Tagging:} We treat the task of valence prediction of the four narratives of a subject as a sequence tagging problem. The intuition behind this approach comes from the fact that the narratives of a subject are collected in a specific temporal order. Our hypothesis is that a sequential model might capture trends in users behavior which could help predict their state-of-mind after the narratives.
 For example, it might capture that a user is talking about his school-life in most of the narratives and may associate this fact with the current valence state.
 
 Specifically, our architecture first encodes the narratives in fixed-length vectors, in a continuous space. Then uni-directional RNN with GRU cells, is used for tagging the sequence of vectors of all narratives. Figure \ref{fig:architecture} provides a visual representation of the architecture. The first layer is an Embedding layer, which retrieves embeddings for the words in the narrative. The embeddings are then fed to a bi-directional RNN (GRU). Next, an attention layer assigns weights to each of the hidden states in the bi-directional RNN, to combine them and generate a vector representation of the narrative. Aftewards, a uni-directional RNN (GRU) layer consumes these narrative encodings as inputs and produces an output at each timestamp, which then is passed through a softmax layer to get the probability of each class. Additional features like \textit{polarity} and \textit{prev valence class} for each narrative can be concatenated with the document encodings. 
 The attention weights can be used to analyze words, phrases and their position in the narratives which are important for the classification.
 The unidirectional RNN ensures that only previous context is considered while predicting valence for a particular narrative. Hence, in this architecture, the first narrative helps in the prediction of all the narratives while the fourth narrative helps only in the prediction of the valence of the last narrative.\\
{\textbf{Context Pair:} To study the effect of immediate previous context on the classification we create a set of pairs of consecutive narratives. We use these pairs as input to predict the valence of the second narrative. To perform this experiment we modify the last (RNN) layer of the above architecture, to predict valence class only for the last timestep. 
In this way, we convert a sequence tagger into a sequence classifier. We compare the results of the two strategies in Section \ref{sect:results}.

\subsection{Preprocessing}
The corpus was released as a part of the COMPARE-2018 challenge for the task of valence prediction. The objective of the competition was to predict the valence class, given 8 seconds segments of the recordings of a narrative. In this way from the initial corpus of 4 narratives given by the 100 participants, 2313 fragments were extracted as train/development/test data (846/742/725) \cite{schuller2018interspeech}. 
Participants to the competition had, therefore, access to the acoustics of the data, but only to fragments of the current narrative.

Our work, on the other hand, focuses on exploring current narratives in their full length and previously recounted narratives by the same individual for valence prediction. Compared to the challenge, thus, the size of our source data is 400 samples (100 subjects, 4 narratives each) before preprocessing.
Another difference compared to the challenge is that we decided not to utilize the audio data and use as our input only the manual text transcriptions of the speech data. Transcriptions were performed by multiple transcribers, resulting in inconsistent formats. The inconsistencies are found in the usage of punctuations, capitalization of words, sentence segmentation and handling of disfluencies. In order to make the data consistent, we perform several preprocessing steps, including punctuation removal and conversion to lower case.

Another important preprocessing step involves removal of some samples from the data. In the challenge, some narratives were rejected as there were some issues with the speech files. It did not affect the challenge as they consider the narratives in isolation, without considering the context. In our experiments, we consider only those users for which all four narratives are present since our goal is to study how the previous context of the subject helps improve the valence prediction at any stage. We reject 28 users who meet this criterion, leaving 72 users' data (288 narratives) for experiments. 
For the same reason, we do not evaluate our models on the first narratives of subjects (N1) as they have no previous narrative, although these are used as context in the models.
\begin{table}[htpb]
\centering
\setlength{\tabcolsep}{3pt}
\hspace*{-0.4cm}
\begin{tabular}{l|l|l|l}
\hline
\textbf{\begin{tabular}[c]{@{}l@{}}Model\\ \end{tabular}} & \textbf{\begin{tabular}[c]{@{}l@{}}Narratives\\ used\end{tabular}} & \textbf{Features} &
\textbf{\begin{tabular}[c]{@{}l@{}}Accuracy\end{tabular}} \\ \hline
\hline
\multirow{4}{*}{linear SVM} & $N_{t}$ & $\mu$ word emb & 55.5 $\pm$5.0 \\ \cline{2-4} 
 & $N_{t}$ & $\mu$ word emb, pol & 57.8 $\pm$4.8 \\ \cline{2-4} 
 & $N_{t}$ & tf-idf, pol & 57.8 $\pm$4.5 \\ \cline{2-4} 
 & $N_{t-1}$, $N_{t}$ & tf-idf, pol & 59.7 $\pm$5.9 \\ \hline \hline
biRNN + attn & $N_{t}$ & word emb, pol & 58.2 $\pm$6.8 \\ \hline
\begin{tabular}[c]{@{}l@{}}encoder\\(biRNN + attn) \\+ RNN\\(context pair)\end{tabular} & $N_{t-1}$, $N_{t}$ & word emb, pol & \textbf{62.4} $\pm$8.7 \\ \hline
\begin{tabular}[c]{@{}l@{}}encoder\\(biRNN + attn) \\+ RNN\\(sequence tagging)\end{tabular} & $N_{0},...,N_{t}$ & word emb, pol & 61.8 $\pm$6.4 \\ \hline
\end{tabular}
\caption{We report Accuracy (with standard deviation) for our models on valence prediction. All experiments were conducted with 5--fold cross--validation.}
\label{tab:results}
\vspace{-0.5cm}
\end{table}
\section{Experiments and Results}
\label{sect:results}
\begin{figure*}[ht]
  \includegraphics[width=\textwidth]{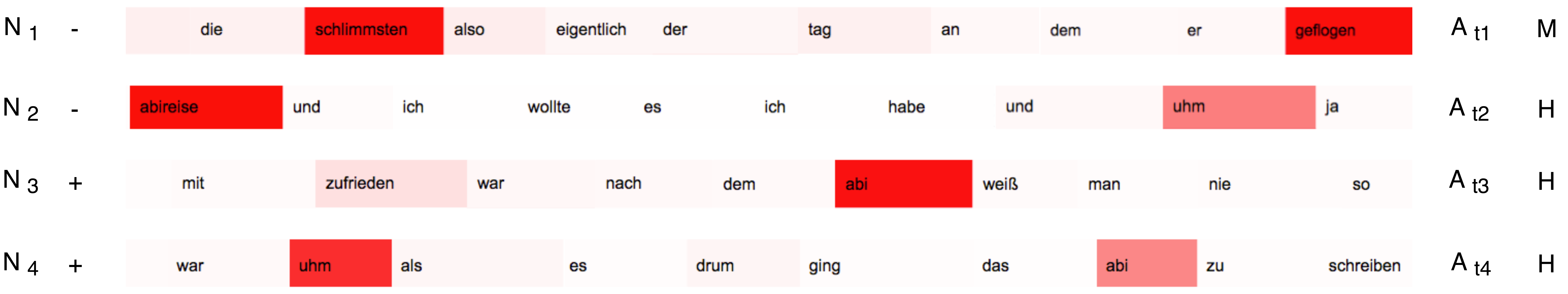}
  \caption{Distribution of attention weights (darker shade of red = higher weight) on four (fragments of) consecutive narratives (N), two negative (-) and two positive (+) by the same individual, in the sequence tagging architecture. The gold valence scores at each timestep of self-reported affect (A) were first medium and then high (M,H,H,H). All scores were correctly predicted by the model.}
  \label{fig:attention}
  \vspace{-0.3cm}
\end{figure*}
In all experiments we use 5-fold cross-validation, ensuring there was no overlap in subjects across the training and validation sets. 
We chose K-fold cross-validation as it allows to use the entire data for training as well as testing, which was useful given the small size of our corpus.

We perform various experiments using different combinations of features described in Section \ref{sect:fatures} and a model from those described in Section \ref{sect:models}. All the main experiments and their results are listed in Table \ref{tab:results}. The first column provides information about the model used. 
In the second column (`Narratives used') we specify the number of narratives utilized by the model for prediction. The context could be the current narrative ($N_{t}$), immediately previous narrative ($N_{t-1}$) or the entire sequence of narratives ($N_{0},...,N_{t}$). The third column, `features' lists the features being extracted and used for the narratives. The last column provides the average accuracy score (with standard deviation) across the 5 folds. 
\subsection{Experiments with Linear SVM:}
In the first set of experiments we use \textit{average word embeddings}
to generate document vectors. In the first experiment we set up a baseline with an accuracy of 55.5\%. In order to verify whether polarity scores calculated using the textblob-de could be helpful, we add it as a feature to the document vector, in the second experiment. We see an increment of around 2.2\% in the accuracy score, to obtain a score of 57.8\%. Polarity score has shown to improve the model performance in many other experiments as well, thus in all further experiments, we use the polarity score as an additional feature.\\
In the second set of experiments, we use tf-idf 
to generate vectors for narratives. Once again, to set up a baseline for this set of experiments, we consider the narrative in isolation and use the tf-idf vector along with the polarity score. We obtain an accuracy score of 57.8\%.\\
To test our intuition that the user context may provide important information for the current valence prediction, in the next experiment, we use the `prev\_val' feature as described in Section \ref{sect:fatures}. 
Since subjects reported valence using a 10-point Likert scale, and these values were not subject-normalized, different people might interpret the scale differently. This experiment is thus aimed at verifying whether the contextual information about previous states-of-mind of the same user could be helpful in predicting the current one.
With this additional feature, we get an accuracy score of 66.3\%, almost 9\% increment from the baseline. This result shows that the context is indeed an important factor for current state-of-mind and previous valence state captures this context precisely. This provides motivation to perform experiments trying to capture context information from the text.\\
In the next experiment we try to add context information from the immediate previous narrative, by concatenating the tf-idf vectors of both narratives, along with the polarity scores. This results in an accuracy of 59.7\% providing an increment of about 1.9 \% over the baseline. 
We perform these experiments using tf-idf and not the word-embeddings as it can provide more insights into how context features help improve results (see Section \ref{sect:disc} for a discussion).
\subsection{Experiments with DNN architectures:}
In this setting, we use the DNN architectures described in Section \ref{sect:dnn}. Our baseline for this set up is a simple BiRNN with attention architecture to predict the valence class considering the narrative in isolation without any context information.  We also concatenate the polarity score for each narrative to their respective encodings. The accuracy of this model is 58.2\%. Next, we use the `sequence tagging' architecture,
for which we get an accuracy of 61.8\%. An improvement of about 4\% is achieved in this experiment. 
The next experiment is based on the `context-pair' method.
Here we use encodings of the current and previous narrative and polarity scores as features to predict the current valence.
This model performs with an accuracy of 62.4\%.

\section{Discussion}
\label{sect:disc}

In order to get further insight about which contextual features seem to be relevant to predict valence, we utilize the attention weights learned by neural models, especially in the entire sequence tagging architecture, as it has access to the full history of narratives. 
Figure \ref{fig:attention} shows the distribution of weights on a sequence of four narratives when predicting the self-reported valence ($A_{t4}$) after the last narrative ($N_{4}$). Due to the limited available space we show only very small fragments of each narrative . As shown in the figure, we find that not only sentiment carrying words and phrases (e.g. happy \textit{zufrieden}), but also emotion triggering concepts and entities, such as people (e.g. grandfather \textit{opa}), events (e.g. exams \textit{abi}), places (e.g. university \textit{uni})from both current and previous narratives seem to have a relevant role for predicting the individual state-of-mind. 
In the example in Figure \ref{fig:attention} the same concept `abi' (high school graduation exam in Germany) receives attention weights across 3 consecutive narratives (`abireise' literally the trip after the high school graduation exam in $N_{2}$, `abi' in $N_{3}$, $N_{4}$). We also notice how disfluencies (`uhm' in $N_{2}$, $N_{4}$) also seem to play an important role for valence prediction, indicating that the model might also learn about subjects' characteristics in speaking style linked to state-of-mind.
We further verify this intuition by analyzing the feature importance assigned by the SVM model trained during the experiment where we concatenate the two tf-idf vectors and the polarity scores.\\
The top features used by the SVM models according to our analysis also show the importance of events (e.g. \textit{abi} is in the top 10 when appearing in the current narrative and in the top 15 weight when appearing in previous ones), disfluencies (e.g. \textit{uhm} is in the top 10 both when appearing in current and when appearing previous narratives). The full list of top features is not shown due to space limitations.
The word \textit{abi} gets more weight in both the models. In the SVM features, the feature \textit{`abi\_prev'} is at the $15^{th}$ position.

\section{Conclusions}
\label{sect:conc}
The results of the experiments performed support the hypothesis that the context of previous narratives recounted by an individual is useful for predicting the current state-of-the-mind of the subject. 
Furthermore, our qualitative analysis using both visualizations of attention weights for neural models and top features for SVM, highlighted how not only emotion words but also other `emotion triggering' concepts (e.g. events, people) and even disfluencies from previous narratives seem to play a role in predicting individuals' valence.

Due to the limited sample size and the lack of longitudinal data, the results should be interpreted with caution. Replication studies preferably with bigger sample sizes, diversity in relation to demographic variables (such as age, gender, etc.) and repeated measurements are needed. Nevertheless, 
the potential of the use of automatized narrative analysis seems promising for future application in mental health care (e.g assistance in diagnostics, evaluation of therapy outcome etc.).

\section{Acknowledgements}
The research leading to these results has received funding from the European Union – H2020 Programme  under grant agreement 826266: COADAPT.
\balance

\bibliographystyle{IEEEtran}

\bibliography{template}


\end{document}